\documentclass[11pt]{article}
\usepackage{graphicx}
\usepackage{longtable,lscape}
\usepackage{float}
\usepackage{url}
\mathchardef\mhyphen="2D

\newcommand{\captionfonts}{\footnotesize}
\makeatletter  
\long\def\@makecaption#1#2{%
  \vskip\abovecaptionskip
  \sbox\@tempboxa{{\captionfonts #1: #2}}%
  \ifdim \wd\@tempboxa >\hsize
    {\captionfonts #1: #2\par}
  \else
    \hbox to\hsize{\hfil\box\@tempboxa\hfil}%
  \fi
  \vskip\belowcaptionskip}
\makeatother   
\begin{document}
\title{The Pet-Fish problem on the World-Wide Web}
\author{Diederik Aerts\\
        \normalsize\itshape
        Center Leo Apostel for Interdisciplinary Studies \\
         \normalsize\itshape
        Vrije Universiteit Brussel, 1160 Brussels, 
       Belgium \\
        \normalsize
        Email: \textsf{diraerts@vub.ac.be} \\ \\
        Marek Czachor\\
        \normalsize\itshape
        Department of Physics \\
        \normalsize\itshape
        Gdansk University of Technology, 80-952 Gdansk, Poland \\
        \normalsize
        Email: \textsf{mczachor@pg.gda.pl} \\ \\
        Bart D'Hooghe\\
        \normalsize\itshape
        Center Leo Apostel for Interdisciplinary Studies \\
         \normalsize\itshape
        Vrije Universiteit Brussel, 1160 Brussels, 
       Belgium \\
        \normalsize
        Email: \textsf{bdhooghe@vub.ac.be} \\ \\
        Sandro Sozzo \\
        \normalsize\itshape
        Department of Physics and INFN \\
         \normalsize\itshape
        University of Salento, 73100 Lecce, Italy \\
        \normalsize
        Email: \textsf{Sandro.Sozzo@le.infn.it}
        }
\date{}
\maketitle
\begin{abstract}
\noindent We identify the presence of Pet-Fish problem situations and the corresponding Guppy effect of concept theory on the World-Wide Web. For this purpose, we introduce absolute weights for words expressing concepts and relative weights between words expressing concepts, and the notion of `meaning bound' between two words expressing concepts, making explicit use of the conceptual structure of the World-Wide Web. The Pet-Fish problem occurs whenever there are exemplars -- in the case of {\it Pet} and {\it Fish} these can be {\it Guppy} or {\it Goldfish} -- for which the meaning bound with respect to the conjunction is stronger than the meaning bounds with respect to the individual concepts.
\end{abstract}

\section{Introduction}
In Aerts \& Gabora (2005a,b), we introduced a modeling scheme for concepts and their combinations that makes use of the mathematical formalism of quantum physics. This quantum modeling scheme has been further worked out in Aerts (2009a) and Aerts (2010a,b). The experimental data we used to create our modeling scheme were data collected in experiments with human subjects that were conducted within the framework of concepts research in psychology (Hampton 1988a,b). These experiments required human subjects to estimate typicalities of exemplars of concepts and their combinations. The results of these estimations were in conflict with how combinations of concepts such as `conjunction' and `disjunction' were expected to behave classically, as prescribed by classical logic or set theory. Hampton called these deviations from classical behavior `overextension' and `underextension', depending on their relation to the classically expected values of typicality (Hampton 1988a,b).

Hampton's experiments, and his search for over and underextended behavior of concepts and their conjunctions and disjunctions, were inspired by the well-known Pet-Fish problem in concept theory, the first discussion of such a deviation from classicality for the conjunction of concepts. Osherson and Smith (1981) considered the conceptual combination of {\it Pet} and {\it Fish} into the conjunction {\it Pet-Fish}, and measured the typicality of different exemplars with respect to the concepts {\it Pet}, {\it Fish} and their conjunction {\it Pet-Fish}. They observed that exemplars such as {\it Guppy} and {\it Goldfish} give rise to a typicality with respect to the conjunction {\it Pet-Fish} which is much bigger than would be expected if the conjunction of {\it Pet} and {\it Fish} were treated from the perspective of classical logic and set theory. Indeed, from a classical perspective, i.e. when modeled by fuzzy set theory applying the maximum rule for conjunction, the typicality of an exemplar with respect to the conjunction of two concepts should not exceed the maximum of the typicalities of this exemplar with respect to both concepts apart. The effect present in the Pet-Fish problem is often referred to as the Guppy effect.

In Aerts and Gabora (2005a,b), we analyzed the Pet-Fish problem in detail, showing how, within the quantum modeling scheme we proposed, the experimental data violating the classically expected values could be modeled correctly. The aim of the present article is to show how the phenomenon underlying the Pet-Fish problem can be identified on the World-Wide Web. More specifically, we can show that the same kind of violation of classicality put forward in the Pet-Fish problem by Osherson and Smith (1981) also occurs when conceptual data are gathered on the World-Wide Web in a specific way. We will explain this below. This way of extracting conceptual structure from the World-Wide Web was put forward recently by one of the authors as part of the elaboration of a new interpretation of quantum mechanics analyzing the violation of Bell inequalities (Aerts 2009b, 2010c). It is currently being elaborated into a quantum mechanical model for the World-Wide Web (Aerts 2010e,f). The present article will show that the technique put forward in Aerts (2009b) and Aerts (2010c,e,f) can also be used to identify the presence of Pet-Fish problem situations on the World-Wide Web.

In recent years, the quantum modeling of situations in cognition and in other domains different from the micro-world has become the focus of research of several scientists working in the newly emerging field of research called `Quantum Interaction'. This has already led to the organization of three successful international workshops (Bruza et al. 2007, 2008, 2009). Several effects observed in the field of cognition have been linked to the presence of quantum structure, more specifically `the disjunction effect' (Busemeyer, Wang and Townsend 2006; Pothos and Busemeyer 2009), `the conjunction fallacy' (Franco 2009), but also the Allais and Ellsberg paradoxes in economy (Allais 1953; Ellsberg 1961; Aerts and D'Hooghe 2009). In the present paper we confine ourselves to an analysis of the Pet-Fish problem, but in Aerts (2010e,f) we show that the other non-classical effects can be studied as well by collecting data on the World-Wide Web in a similar way as we do here for the Pet-Fish problem. To this end, we used Yahoo's search engine to find the numbers of webpages containing certain terms or words expressing concepts and combinations of concepts. The reason why we preferred Yahoo to Google to compile our statistics, is that we found Yahoo to produce more consistent results over time. Since these numbers change over time, we should add that we carried out the web searches necessary for this research on May 4, 2010. 

\section{Relative Weights}

We found 1,290,000,000 webpages containing the word {\it Pet}, 1,100,000,000 webpages containing the word {\it Fish}, and 1,760,000 webpages containing the combination of words {\it Pet-Fish}. Our aim was to find only results containing the exact combination, so that we entered the search term ``pet fish". We then considered another word, {\it Hierarchy} -- we will make clear in the following why we chose such a rather uncommon word like {\it Hierarchy} to start our analysis. We found 4,210,000 webpages containing both words {\it Pet} and {\it Hierarchy}, 6,550,000 webpages containing both words {\it Fish} and {\it Hierarchy} and 1,410 webpages containing both the word expressing the conjunction concept {\it Pet-Fish} and the word {\it Hierarchy}. This means that if we calculate the relative weights, i.e. the number of webpages containing {\it Pet} and {\it Hierarchy} divided by the number of webpages containing {\it Pet}, denoting these relative weights as $w(\mathit{Pet,Hierarchy})$, and equally so for {\it Fish} and {\it Hierarchy}, and {\it Pet-Fish} and {\it Hierarchy}, we find that $w(\mathit{Pet,Hierarchy})=0.00326$, $w(\mathit{Fish,Hierarchy})=0.00595$ and $w(\mathit{Pet\mhyphen Fish,Hierarchy})=0.00080$, respectively. This means that
\begin{eqnarray} \label{classical01}
&w(\mathit{Pet\mhyphen Fish,Hierarchy}) \le w(\mathit{Pet,Hierarchy}) \\ \label{classical02}
&w(\mathit{Pet\mhyphen Fish,Hierarchy}) \le w(\mathit{Fish,Hierarchy})
\end{eqnarray}
Inequalities (\ref{classical01}) and (\ref{classical02}) show that the words {\it Pet}, {\it Fish} and {\it Pet-Fish} behave classically with respect to the word {\it Hierarchy}, if we regard the relative weights as adequate measures for the typicality, within a classical fuzzy set theoretic modeling and apply the maximum rule for the conjunction. In general terms, we therefore define the relative weight $w(A,B)$ of word $B$ with respect to word $A$ as the number of webpages containing both word $A$ and word $B$ divided by the number of webpages containing word $A$.

Let us now consider the word {\it Guppy} and calculate the relative weights of this word with respect to {\it Pet}, {\it Fish} and {\it Pet-Fish}. We found 3,050,000 webpages containing the word {\it Pet} and the word {\it Guppy}, 4,520,000 webpages containing the word {\it Fish} and the word {\it Guppy}, and 37,900 webpages containing the word {\it Pet-Fish} and the word {\it Guppy}. This gives us $w(\mathit{Pet,Guppy})=0.00236$, $w(\mathit{Fish,Guppy})=0.00411$ and $w(\mathit{Pet\mhyphen Fish,Guppy})=0.02153$. With respect to {\it Guppy}, this gives us a relation between the relative weights that is contrary to the one we have with respect to {\it Hierarchy}, namely
\begin{eqnarray}
&w(\mathit{Pet,Guppy}) \le w(\mathit{Pet\mhyphen Fish,Guppy}) \\
&w(\mathit{Fish,Guppy}) \le w(\mathit{Pet\mhyphen Fish,Guppy})
\end{eqnarray}
In other words, this is a manifest instance of non-classical behavior of the concept expressed by the word {\it Guppy} with respect to the concepts expressed by the words {\it Pet}, {\it Fish} and {\it Pet-Fish}.

\section{Correcting Counts}

Before giving further examples to illustrate the frequent occurrence of the Pet-Fish problem on the World-Wide Web, we should briefly point out a specific problem concerning the numbers of webpages yielded by the Yahoo search engine -- as well as by the Google search engine, for that matter. For example, when investigating whether there was a Guppy effect with respect to the concept {\it World}, we found the number of webpages containing {\it Pet} and {\it World} to be 1,030,000,000. The number of webpages containing {\it Pet} and `not' containing {\it World} was 890,000,000. However, the number of webpages containing the word {\it Pet} was 1,290,000,000, i.e. much lower than the sum of the number of webpages containing {\it Pet} and {\it World} (1,030,000,000) and the number of webpages containing {\it Pet} and `not' containing {\it World} (890,000,000). So the counts by Yahoo -- and equally so by Google -- were incorrect here. We noticed that this error occurred when making combinations with words that are very abundant on the World-Wide Web, such as {\it World}. Indeed, the number of webpages containing {\it World} was 11,100,000,000. To introduce a correction for this error, we proceeded as follows. We divided the number of webpages containing {\it Pet}, i.e. 1,290,000,000, by the the sum of the number of webpages containing {\it Pet} and {\it World} and the number of webpages containing {\it Pet} and `not' containing {\it World}, i.e. 1,030,000,000 + 890,000,000, which gave a correction factor 0.671875. We then multiplied this correction factor by the number of webpages containing {\it Pet} and {\it Word} found using Yahoo, which gave us 692,031,250. We consider this `corrected number' to be a fair estimate of the number of webpages containing both words {\it Pet} and {\it World}.

\section{The Guppy Effect and Meaning}

We gathered data on the World-Wide Web for several concepts with the aim of understanding the Guppy effect, and the results are listed in Table 1. {\it Tot. N} is the total number of webpages found in our searches conducted on May 4 using the Yahoo search engine. {\it Rel. N} is the relative number of webpages, e.g. in the first column it is the number of webpages containing {\it Pet} and one of the other words considered, viz. {\it Guppy}, {\it World}, {\it Spelling}, {\it House}, {\it Goldfish} and {\it Hierarchy}. In the second column. {\it Rel. N} is the number of webpages containing {\it Fish} and one of the other aforementioned words, while in the third column, it is the number of webpages containing {\it Pet-Fish} and one of the other words. {\it Rel. -N} is the relative number of webpages `not' containing the word concerned. For example, in the region where {\it Guppy} is considered, in the first column, {\it Rel. N} is the number of webpages containing {\it Pet} and `not' containing {\it Guppy}. In the second column of the region where {\it Guppy} is considered, it is the number of webpages containing {\it Fish} and `not' containing {\it Guppy}, and in the third column of this region, it is the number of webpages containing {\it Pet-Fish} and `not' containing {\it Guppy}. An analogous approach applies to the other regions, considering the words {\it World}, {\it Spelling}, {\it House}, {\it Goldfish} and {\it Hierarchy}. {\it Corr.} is the `correction factor' for every region and for every column within each region. For example, for the first column, the number of webpages containing {\it Pet} divided by the sum of the number of webpages {\it Rel. N} and {\it Rel -N} of the region considered, {\it Rel. N corr.} is the corrected relative number of webpages, i.e. in each column of every region of a specific word it is the {\it Rel N} of this region multiplied by the correction factor of this region and the column concerned. Finally, {\it Rel. w} is the relative weight, which is the value that for each region corresponding to a specific word is given by the expression {\it Rel. N corr.} divided by the {\it Tot. N}.

\begin{table}[htdp]
\caption{Data collected on the World-Wide Web illustrating the Pet-Fish problem of concept research in psychology}
\medskip
\begin{center}
\footnotesize
\begin{tabular}{|c|c|c|c|}
\hline
& {\it Pet} & {\it Fish} & {\it Pet-Fish} \\
\hline
{\it Tot. N} & 1,290,000,000 & 1,100,000,000 & 1,760,000 \\
\hline
\multicolumn{4}{|l|}{\it Guppy} \\
\hline
{\it Tot. N} & \multicolumn{3}{|l|}{12,900,000} \\
{\it Abs. w} & \multicolumn{3}{|l|}{0.000234545} \\
\hline
{\it Rel. N} & 3,050,000 & 4,520,000 & 37,900 \\
{\it Rel. -N} & 1,290,000,000 & 1,100,000,000 & 1,710,000 \\
{\it Corr.} & 0.9976412 & 0.9959077 & 1.0069226 \\
{\it Rel. N corr.} & 3,042,806 & 4,501,503 & 38,162 \\
{\it Rel. w} & 0.0023588 & 0.0040923 & 0.0216832 \\
{\it M} & 10.0567 & 17.4477 & 92.4476 \\
\hline
\multicolumn{4}{|l|}{\it World} \\
\hline
{\it Tot. N} & \multicolumn{3}{|l|}{11,100,000,000} \\
{\it Abs. w} & \multicolumn{3}{|l|}{0.201818182} \\
\hline
{\it Rel. N} & 1,030,000,000 & 719,000,000 & 737,000 \\
{\it Rel. -N} & 890,000,000 & 633,000,000 & 970,000 \\
{\it Corr.} & 0.671875 & 0.8136095 & 1.0310486 \\
{\it Rel. N corr.} & 692,031,250 & 584,985,207 & 759,882 \\
{\it Rel. w} & 0.5364583 & 0.5318047 & 0.4317516 \\
{\it M} & 2.6581 & 2.6351 & 2.1393 \\
\hline
\multicolumn{4}{|l|}{\it Spelling} \\
\hline
{\it Tot. N} & \multicolumn{3}{|l|}{291,000,000} \\
{\it Abs. w} & \multicolumn{3}{|l|}{0.005290909} \\
\hline
{\it Rel. N} & 32,100,000 & 29,000,000 & 40,200 \\
{\it Rel. -N} & 1,280,000,000 & 1,090,000,000 & 124,000,000 \\
{\it Corr.} & 0.9831568 & 0.9830206 & 0.9998864 \\
{\it Rel. N corr.} & 31,559,332 & 28,507,596 & 40,195 \\
{\it Rel. w} & 0.0244646 & 0.0259160 & 0.0228383 \\
{\it M} & 4.6239 & 4.8982 & 4.3165 \\
\hline
\multicolumn{4}{|l|}{\it House} \\
\hline
{\it Tot. N} & \multicolumn{3}{|l|}{4,880,000,000} \\
{\it Abs. w} & \multicolumn{3}{|l|}{0.088727273} \\
\hline
{\it Rel. N} & 683,000,000 & 316,000,000 & 431,000 \\
{\it Rel. -N} & 1,020,000,000 & 1,280,000,000 & 1,500,000 \\
{\it Corr.} & 0.7574868 & 0.6892231 & 0.9114448 \\
{\it Rel. N corr.} & 517,363,476 & 217,794,486 & 392,832 \\
{\it Rel. w} & 0.4010570 & 0.1979950 & 0.2232004 \\
{\it M} & 4.5201 & 2.2315 & 2.5156 \\
\hline
\multicolumn{4}{|l|}{\it Goldfish} \\
\hline
{\it Tot. N} & \multicolumn{3}{|l|}{32,500,000} \\
{\it Abs. w} & \multicolumn{3}{|l|}{0.000590909} \\
\hline
{\it Rel. N} & 9,790,000 & 9,790,000 & 225,000 \\
{\it Rel. -N} & 1,280,000,000 & 1,280,000,000 & 1,500,000 \\
{\it Corr.} & 1.0001628 & 0.8528520 & 1.0202899 \\
{\it Rel. N corr.} & 9,791,593 & 8,349,421 & 229,565 \\
{\it Rel. w} & 0.0075904 & 0.0075904 & 0.1304348 \\
{\it M} & 12.8453 & 12.8453 & 220.7358 \\
\hline
\multicolumn{4}{|l|}{\it Hierarchy} \\
\hline
{\it Tot. N} & \multicolumn{3}{|l|}{79,200,000} \\
{\it Abs. w} & \multicolumn{3}{|l|}{0.00144} \\
\hline
{\it Rel. N} & 4,210,000 & 6,550,000 & 1,410 \\
{\it Rel. -N} & 1,290,000,000 & 1,090,000,000 & 1,760,000 \\
{\it Corr.} & 0.996747 & 1.0031462 & 0.9991995 \\
{\it Rel. N corr.} & 4,196,305 & 6,570,607 & 1,408 \\
{\it Rel. w} & 0.003253 & 0.005973 & 0.0008005 \\
{\it M} & 2.2590 & 4.1481 & 0.5559 \\
\hline
\end{tabular}
\end{center}
\label{default}
\end{table}%

\normalsize
The relative weight is a number between 0 and 1. For a given region and for the first column, it indicates the fraction of webpages containing {\it Pet} and the word corresponding to this region with respect to the total number of webpages containing {\it Pet}.
For the second and third columns, it indicates these same fractions with respect to {\it Fish} and {\it Pet-Fish}.

Before we interpret the results of the data we collected on the World-Wide Web, we should explain yet another aspect of our analysis. There is hardly any in-depth knowledge about `how many webpages in all the World-Wide Web comprises at this moment'. Estimations of how large this total number may be vary because the outcome somehow depends on such factors as the manner of counting and the type of pages considered. For example, when we entered the word `and' in the Yahoo search engine, we found 34,900,000,000 webpages. When we entered the word `the', it returned 36,400,000,000 hits, while the number `1' was found 49,000,000,000 times. Most probably, the latter search yielded the largest number of hits we were able to get in this way, since `1' also counts webpages in languages different from English, unlike the cases of `the' and `and'. The reason why we would like to know the total number of webpages is that this allows us to introduce a quantity with respect to any concept. We have called this quantity the `absolute weight' of this concept, i.e. the number of webpages containing the word expressing this concept divided by the totality of webpages comprising the World-Wide Web as a whole. Table 1 lists the results of our calculations of this quantity for the different concepts considered, assuming 55,000,000,000 to be the total number of webpages of the World-Wide Web indexed by Yahoo (Kunder 2010). 

We will now interpret our results. For each concept, the absolute weight provides a measure of its overall presence on the World-Wide Web. {\it Guppy} and {\it Goldfish} turn out to have the lowest absolute weights of all the concepts we considered, viz. $0.000234545=2.34545 \cdot 10^{-4}$ and $0.000590909=5.90909 \cdot 10^{-4}$, respectively. They are followed by {\it Spelling}, with an absolute weight of $0.005290909=5.290909 \cdot 10^{-3}$, and by {\it House} and {\it Hierarchy}, with absolute weights of $0.088727273=0.88727273 \cdot 10^{-2}$ and $0.088727273=0.88727273 \cdot 10^{-2}$, respectively. The word {\it World} has an absolute weight of $0.201818182=2.01818182 \cdot 10^{-1}$. For each of the three words {\it Pet}, {\it Fish} and {\it Pet-Fish}, the corresponding relative weights provide the means to measure their presence in `meaning contexts'. Now, if the measure of the presence of a word $B$ in the `meaning context' of another word $A$ equals the measure of the overall presence of this word $B$, this would indicate that there is no specific `meaning bound' between $A$ and $B$. Hence, if we divide the relative weight of word $B$ with respect to $A$ by the absolute weight of word $B$, the resulting number is capable of expressing in quantitive terms the meaning bound between the words $A$ and $B$. This number will be equal to 1 if there is no meaning bound, i.e. no meaning bound stronger than the overall meaning bound related to the fact that words are members of the whole World-Wide Web. It will be larger than 1 if $A$ and $B$ have -- what we have called -- an `attractive meaning bound' and it will be smaller than 1 if $A$ and $B$ have Ð what we have called Ð a `repulsive meaning bound'. We have calculated these quantities and called them $M(A,B)$ for the words expressing the concepts which we considered with respect to the Pet-Fish problem. They are given in Table 1. Let us remark that
\begin{equation}
M(A,B)={n(A,B)/n(A) \over n(B)/n(\mathrm{www})}={n(A,B)n(\mathrm{www}) \over n(B)n(A)}
\end{equation}
where $n(A)$ and $n(B)$ are the number of webpages containing word $A$ and word $B$, respectively, $n(A,B)$ is the number of webpages containing word $A$ and word $B$, and $n(\mathrm{www})$ is the total number of webpages of the World-Wide Web. It is interesting to note that $M(A,B)=M(B,A)$, and hence $M(A,B)$ is symmetric in $A$ and $B$, which means that we can really speak of a `meaning bound between word $A$ and word $B$'.

So we can see in Table 1 that the meaning bounds between {\it Guppy} and {\it Pet}, {\it Fish} and {\it Pet-Fish} are 10, 17 and 92, respectively. These are all three strong attractive meaning bounds, but the meaning bound with respect to {\it Pet-Fish} is much bigger than the meaning bounds with respect to {\it Pet} and {\it Fish}. Of the words considered, only {\it Goldfish} gives rise to an even more strongly pronounced effect of attractive meaning bound, namely 12, 12 and 220, respectively, with respect to {\it Pet}, {\it Fish} and {\it Pet-Fish}. All of the meaning bounds we have calculated are attractive, except the one of {\it Hierarchy} with respect to {\it Pet-Fish}, which is repulsive. This means that the relative occurrence of the word {\it Hierarchy} in webpages containing the concept {\it Pet-Fish} is smaller than its overall occurrence on the overall World-Wide Web.

To explain the Guppy effect on the World-Wide Web, we want to put forward an approach based on this meaning bound. For a conjunction of concepts, there may be exemplars of the concepts that have stronger attractive meaning bounds with respect to the conjunction than their meaning bounds with respect to each of the individual concepts, and, if this is the case, the Guppy effect appears. If we look again at the examples given in Table 1, we can see that different situations are possible. For the words {\it World} and {\it Spelling}, the meaning bound with respect to {\it Pet Fish} is less attractive than it is with respect to {\it Pet} and to {\it Fish}. And indeed, as we mentioned already, for the word {\it Hierarchy}, the meaning bound with respect to {\it Pet-Fish} is repulsive, while it is attractive with respect to {\it Pet} and {\it Fish}. For the word {\it House}, the meaning bound with respect to {\it Pet-Fish} is less attractive than with respect to {\it Pet}, but more attractive than with respect to {\it Fish}.

\section{Conclusion}
Although not completely equivalent, the Guppy effect we have identified on the World-Wide Web is strongly related to the one identified in concept research with respect to typicality (Osherson and Smith 1981) and membership weight (Hampton 1988a). We also are quite convinced that both effects have the same origin, and that this origin is in some sense revealed more clearly on the World-Wide Web than it is within the context of the original psychological experiments. The origin is that `the extent to which exemplars are present in the meaning landscape surrounding concepts and their conjunctions' does not follow the inequalities which are classically supposed to be fulfilled if typicality and membership are looked upon from a fuzzy set theoretic perspective, as in the original analyses put forward by Osherson and Smith (1981), and by Hampton (1988a). This `extent of presence in the meaning landscape surrounding the concepts considered and their conjunctions' is a factor that is equally determining as those originating in typicality and membership. In other words, when typicality or membership are measured in psychological experiments, they are strongly influenced by the effect of `the extent of presence in the meaning landscape surrounding the concepts considered and their conjunctions'. In our analysis of the Pet-Fish problem on the World-Wide Web, we directly calculated this `extent of presence in the meaning landscape surrounding the concepts considered and their conjunctions'. In Aerts (2010e), we showed how the analysis of the Pet-Fish problem put forward in the present article is a direct consequence of the quantum-mechanical model for the measurement of meaning elaborated in that publication, and how the mathematical expression for the `meaning bound' we introduced here follows from the quantum-mechanical weight structure.

\section{References}
Aerts, D. (2009a). Quantum structure in cognition. {\it Journal of Mathematical Psychology}, {\bf 53}, pp. 314-348.

\smallskip
\noindent
Aerts, D. (2009b). Quantum particles as conceptual entities: A possible explanatory framework for quantum theory. {\it Foundations of Science}, {\bf 14}, pp. 361-411. 

\smallskip
\noindent
Aerts, D. (2010a). Quantum interference and superposition in cognition: Development of a theory for the disjunction of concepts. In D. Aerts, B. D'Hooghe and N. Note (Eds.), {\it Worldviews, Science and Us: Bridging Knowledge and Its Implications for Our Perspectives of the World}. Singapore: World Scientific. 

\smallskip
\noindent
Aerts, D. (2010b). General quantum modeling of combining concepts: A quantum field model in Fock space. {\it Foundations of Science}.

\smallskip
\noindent
Aerts, D. (2010c). Interpreting quantum particles as conceptual entities. {\it International Journal of Theoretical Physics}. 

\smallskip
\noindent
Aerts, D. (2010d). A potentiality and conceptuality interpretation of quantum physics. {\it Philosophica}.

\smallskip
\noindent
Aerts, D. (2010e). Measuring meaning on the World-Wide Web. Preprint Center Leo Apostel. Archive reference and link: http://arxiv.org/abs/1006.1786.

\smallskip
\noindent
Aerts, D. (2010f). The conjunction fallacy on the World-Wide Web. Preprint Center Leo Apostel.

\smallskip
\noindent
Aerts, D. and D'Hooghe, B. (2009). Classical logical versus quantum conceptual thought: Examples in economics, decision theory and concept theory. In P. D. Bruza, D. Sofge, W. Lawless, C. J. van Rijsbergen and M. Klusch (Eds.), {\it Proceedings of QI 2009-Third International Symposium on Quantum Interaction, Book series: Lecture Notes in Computer Science}, {\bf 5494}, pp. 128-142. Berlin, Heidelberg: Springer.

\smallskip
\noindent
Aerts, D. and Gabora, L. (2005a). A theory of concepts and their combinations I: The structure of the sets of contexts and properties. {\it Kybernetes}, {\bf 34}, pp. 167-191.

\smallskip
\noindent
Aerts, D. and Gabora, L. (2005b). A theory of concepts and their combinations II: A Hilbert space representation. {\it Kybernetes}, {\bf 34}, pp. 192-221.

\smallskip
\noindent
Allais, M. (1953). Le comportement de l'homme rationnel devant le risque: Critique des postulats et axiomes de l'Ecole Americaine. {\it Econometrica}, {\bf 21}, pp. 503-546.

\smallskip
\noindent
Bruza, P. D., Lawless, W., van Rijsbergen, C. J. and Sofge, D. (Eds.) (2007). {\it Proceedings of the AAAI spring
symposium on quantum interaction}. AAAI Press. 

\smallskip
\noindent
Bruza, P. D., Lawless, W., van Rijsbergen, C. J. and Sofge, D. (Eds.). (2008). {\it Quantum interaction: Proceedings of the
second quantum interaction symposium}. London: College Publications.

\smallskip
\noindent
Bruza, P. D., Sofge, D., Lawless, W., van Rijsbergen, C. J. and Klusch, M. (Eds.). (2009). {\it Lecture Notes in Artificial
Intelligence}, {\bf 5494}. {\it Proceedings of the Third Quantum Interaction Symposium}. Berlin, Heidelberg: Springer.

\smallskip
\noindent
Busemeyer, J. R., Wang, Z. and Townsend, J. T. (2006). Quantum dynamics of human decision making. {\it Journal of
Mathematical Psychology}, {\bf 50}, pp. 220-241.

\smallskip
\noindent
Ellsberg, D. (1961). Risk, ambiguity, and the Savage axioms. {\it Quarterly Journal of Economics}, {\bf 75}, pp. 643-669.

\smallskip
\noindent
Franco, R. (2009). The conjunctive fallacy and interference effects. {\it Journal of Mathematical Psychology}, {\bf 53}, pp. 415-422.

\smallskip
\noindent
Hampton, J. A. (1988a). Overextension of conjunctive concepts: Evidence for a unitary model for concept typicality and class inclusion. {\it Journal of Experimental Psychology: Learning, Memory, and
Cognition}, {\bf 14}, pp. 12-32.

\smallskip
\noindent
Hampton, J. A. (1988b). Disjunction of natural concepts. {\it Memory \& Cognition}, {\bf 16}, pp. 579-591.

\smallskip
\noindent
Kunder, M. (2010). The size of the World Wide Web. http://www.worldwidewebsize.com

\smallskip
\noindent
Osherson, D. N. and Smith, E. E. (1981). On the adequacy of prototype theory as a theory of concepts. {\it Cognition}, {\bf 9}, pp. 35-58.

\smallskip
\noindent
Pothos, E. M. and Busemeyer, J. R. (2009). A quantum probability explanation for violations of `rational' decision theory. {\it Proceedings of the Royal Society B}.

\end{document}